\documentclass[runningheads]{llncs}
\usepackage[T1]{fontenc}
\usepackage{amsfonts}

\usepackage{graphicx}
\usepackage{tikz}
\usetikzlibrary{trees}
\usepackage{forest}
\usepackage{amsmath}
\usepackage{xcolor}
\usepackage[sectionbib,numbers]{natbib}
\usepackage{multirow}
\usepackage{array}

\newcolumntype{P}[1]{>{\raggedright\arraybackslash}p{#1}}

\begin{document}
\title{Reliable XAI Explanations in Sudden Cardiac Death Prediction for Chagas Cardiomyopathy}
\titlerunning{Reliable XAI Explanations in Sudden Cardiac Death Prediction}
\author{Vinícius P. Chagas\inst{1}\orcidID{0009-0003-3871-640X} \and
Luiz H. T. Viana\inst{1}\orcidID{0009-0005-1509-6038} \and
Mac M. da S. Carlos\inst{1}\orcidID{0009-0000-3528-614X} \and
João P. V. Madeiro\inst{3}\orcidID{0000-0001-6511-6707} \and
Roberto C. Pedrosa\inst{2}\orcidID{0000-0002-3270-1595} \and
Thiago A. Rocha\inst{1}\orcidID{0000-0001-7037-9683} \and
Carlos H. L. Cavalcante\inst{1}\orcidID{0000-0001-9395-8338}
}
\authorrunning{V. Peixoto Chagas et al.}
\institute{Federal Institute of Education and Technology of Ceara, Maracanau, Ceara, Brazil \\
\email{\{vinicius.peixoto.chagas61, henrique.viana06, mac.myller.silva07\}@aluno.ifce.edu.br; \\
\email{\{thiago.alves, henriqueleitao\}@ifce.edu.br}} 
\and
Edson Saad Heart Institute – Federal University of Rio de Janeiro, Rio de Janeiro, Brazil\\
\email{coury@hucff.ufrj.br} 
\and
Universidade Feferal do Ceará (UFC), CE \\
\email{jpaulo.vale@dc.ufc.br}
}
\maketitle % typeset the header of the contribution
\begin{abstract}

Sudden cardiac death (SCD) is unpredictable, and its prediction in Chagas cardiomyopathy (CC) remains a significant challenge, especially in patients not classified as high risk. While AI and machine learning models improve risk stratification, their adoption is hindered by a lack of transparency, as they are often perceived as \textit{black boxes} with unclear decision-making processes. Some approaches apply heuristic explanations without correctness guarantees, leading to mistakes in the decision-making process. 
To address this, we apply a logic-based explainability method with correctness guarantees to the problem of SCD prediction in CC. This explainability method, applied to an AI classifier with over 95\% accuracy and recall, demonstrated strong predictive performance and 100\% explanation fidelity. When compared to state-of-the-art heuristic methods, it showed superior consistency and robustness.
This approach enhances clinical trust, facilitates the integration of AI-driven tools into practice, and promotes large-scale deployment, particularly in endemic regions where it is most needed.

%
% propose a novel approach that integrates explainability methods with correctness guarantees, generating interpretable explanations for an SCD prediction model in CC. 

% Our explanation methods

\sloppy
\keywords{Machine Learning \and Explainable Artificial Intelligence \and Logic-Based Explanations \and Sudden Cardiac Death \and Chagas Disease}
\end{abstract}

\section{Introduction}

Chagas disease (CD), caused by the parasite Trypanosoma cruzi, is endemic in Brazil and Latin America. Its incidence has been rising in North America, Europe, Japan, and Australia, mainly due to migration. It is estimated that between six and seven million individuals across the globe are infected, many of whom risk severe cardiac and digestive complications if untreated \cite{who_chagas_disease}. 

Sudden cardiac death (SCD) is a significant concern in the context of CD, accounting for approximately 45\% of deaths among patients with Chagas cardiomyopathy (CC) \cite{perez2018chagas}. Its occurrence rate (2.4\% per year) surpasses that of the general population \cite{rassi2013another}, highlighting the severe cardiac risks associated with the disease. The unpredictability of SCD, both in general and in the specific context of CD, is widely documented in the literature \cite{malik2015chagas}, \cite{perez2018chagas}, and \cite{leoni2023bridging}, which has driven several studies aimed at its prediction.

In the specific context of CC, several studies have shown that machine learning (ML) algorithms can classify patients with CD with the same or higher precision than physicians \cite{barker2022machine}. Regarding the prediction of SCD in CD, other studies have employed artificial intelligence (AI) for this purpose \cite{alberto2020association, cavalcante2023sudden, pedrosa2024risk}.
The lack of transparency in these models often leads to distrust among clinicians \cite{daneshjou2021lack}. Explainable Artificial Intelligence (XAI) emerges as a solution, making AI models transparent and understandable. XAI models can provide doctors with an understanding of how the system operates, which data were used, and which factors influenced their conclusions \cite{lotsch2021explainable}. However, commonly used XAI models like LIME~\cite{limeribeiro2016should} or Anchors~\cite{ribeiro2018anchors} do not provide correctness guarantees for their explanations. This means that two different input instances, each associated with a different output class, can have the same explanation. This lack of fidelity may lead to a misunderstanding of the model \cite{gosiewska2019not}.
%This means that an explanation for an instance can match the data of another that is classified differently by the model. 
 
In contrast, logic-based abductive explanations aim to identify non-redudant sets of conditions that are sufficient to guarantee the prediction and offer guarantees of correctness \cite{ignatiev2019validating}. While such methods have been proposed in the literature, they remain largely underutilized in clinical applications. This is particularly true in the context of cardiovascular risk prediction for neglected diseases such as Chagas disease (CD), where explainability can play a crucial role in clinical adoption \cite{ennab2024enhancing}. Developing explainable systems to support the diagnosis of sudden cardiac death (SCD) risk in Chagas disease (CD) patients can reduce uncertainty and foster greater trust among medical professionals. 

In this work, we investigate the application of logic-based abductive explanations to the prediction of sudden cardiac death (SCD) in patients with Chagas cardiomyopathy. Our contribution lies in adapting these methods to a high-stakes medical scenario, demonstrating their practical utility in producing trustworthy and human-interpretable explanations. We further provide a comparative empirical evaluation against widely used heuristic XAI methods, assessing fidelity, running time, and explanation size.

To evaluate our approach, we use real-world clinical data collected over three decades from patients diagnosed with Chagas disease in Brazil. The dataset comprises a limited number of samples, reflecting the practical challenges of studying neglected tropical diseases (NTDs) such as Chagas, including underreporting, limited funding, and scarcity of large-scale structured databases. Given the tabular structure and relatively small size of the data, we adopted XGBoost \cite{chen2016xgboost} as our predictive model due to its high performance and robustness in such conditions \cite{uddin2024confirming}. Additionally, logic-based methods have been shown to efficiently generate explanations for XGBoost models, making them a practical choice for real-world applications \cite{Xreason2022}.

%Our approach achieved high predictive performance, with a recall of 95.00\% and an AUC of 95.00\%, while maintaining 100\% fidelity in its explanations. 

The trained XGBoost model achieved high predictive performance, with a recall of 95.00\% and an AUC of 95.00\%. The explanations generated maintained 100\% fidelity with respect to the model. These results indicate that our approach is suitable for integration into clinical decision support (CDS) systems, where reliability and transparency are critical for adoption by healthcare professionals \cite{shakibaei2024influential}.

%Given the tabular nature and limited size of the dataset employed in this study, tree-based models are especially appropriate, as they tend to deliver strong predictive performance under these conditions \cite{uddin2024confirming}.

%We extract decision paths from the XGBoost model’s predictions and input them into an adaptable explainable AI (XAI) framework, which enhances interpretability. This method provides explanations with guaranteed correctness for the model's predictions.

This study consists of five sections. The first introduces the research problem. The second reviews essential background concepts for understanding the work. The third outlines the methodology, describing the proposed model and system architecture. The fourth presents the experimental results. The final section concludes with reflections on the model and suggestions for future work.

\section{Background}

This section aims to present basic concepts necessary for a general understanding of the work.

\subsection{Sudden cardiac death (SCD)}
Sudden cardiac death (SCD) refers to abrupt heart malfunction resulting in instantaneous collapse, which needs suspension of the heart-lung machine and immediate medical attention, e.g., cardiopulmonary resuscitation, to normalize the blood circulation. A heart is diagnosed with the cardiac origin mechanism when only other causes like cerebrovascular events, acute systemic illness, drug overdose, metabolic disturbances, and terminal conditions are ruled out. resulting in collapse and requiring immediate interventions, such as cardiopulmonary resuscitation, to restore circulation \cite{hinkle1982clinical}. A cardiac origin is diagnosed only after excluding other potential causes, including non-cardiac vascular events, acute systemic illnesses, drug overdoses, metabolic imbalances, or end-stage diseases \cite{hinkle1982clinical}.
Determining the likelihood of SCD is even more complicated due to problems with data collection, such as the sources of data being unavailable and the low incidence but high magnitude of the events \cite{barkauskas2024rare}.

\subsection{Gradient Boosted Trees}

Although several machine learning techniques have been effectively used on medical datasets \cite{cavalcante2023sudden, barker2022machine}, their interpretability can be difficult. As a result, researchers are increasingly focusing on developing techniques that not only achieve high predictive performance but also improve the transparency and understanding of how these algorithms make predictions \cite{ignatiev2019validating}.

Considering the tabular structure and relatively small size of the dataset used in this study, tree-based models are especially appropriate \cite{uddin2024confirming} in terms of learning potential. Moreover, their structural properties make them directly compatible with logic-based explanation methods, enabling formal and reliable reasoning about predictions.

\begin{figure}
  \centering
  %\tiny
  \begin{forest}
    for tree={fill=white, draw, text centered, font=\sffamily, edge={thick, -{[]}, draw}}
      [Is Age < 30?
        [Is Income < 50k?
          [Output: -1]
          [Output: 1]
        ]
        [Is Age < 25?
          [Output: 3]
          [Output: -2]
        ]
      ]
  \end{forest}
  \caption{Example of a regression tree in XGBoost (left is the path for False and right is the path for True)}
  \label{fig:decision_tree}
\end{figure}
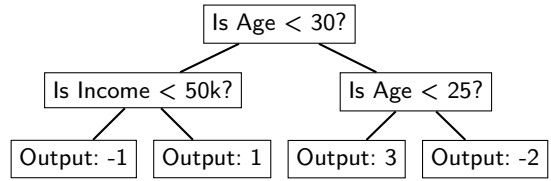

With this context, we employed XGBoost\footnote{https://github.com/dmlc/xgboost}, a machine learning algorithm that implements the gradient boosting framework using regression trees. It is a widely adopted, gradient-boosted tree algorithm that incorporates several performance optimizations\cite{chen2016xgboost}. In binary classification, XGBoost constructs an ensemble of regression trees during training, each producing an output value. Figure~\ref{fig:decision_tree} represents an example regression tree used by the XGBoost classifier. 

The final prediction is the sum of the outputs of all trees, where a positive sum indicates class 1 and a negative sum indicates class 0. Additionally, XGBoost enables the extraction of trained trees, allowing their representation as logical formulas. This capability supports the application of logic-based methods to derive explanations with formal guarantees of correctness.

%it meets the need for logic-based explanations through the \textit{booster} attribute, which allows the extraction of trained trees in textual or tabular formats. This feature provides direct access to the model's internal structure, supporting detailed interpretation and analysis.
% while maintaining the core structure of regression trees

\subsection{First-order Logic over LRA}\label{subsec:logic}

In this work, we use first-order logic (FOL) to generate explanations with guarantees of correctness. We use quantifier-free first-order formulas over the theory of linear real arithmetic (LRA) \cite{kroening2016decision}. Variables are allowed to take values from the real numbers $\mathbb{R}$. We consider formulas as $\sum^n_{i=1} w_i f_i \leq b$, such that $n \geq 1$ is a natural number, each $w_i$ and $b$ are fixed real numbers, and each $f_i$ is a first-order variable. As usual, if $F$ and $G$ are formulas, then $(F \wedge G), (F \vee G), (\neg F), (F \to G)$ are formulas. We allow the use of other letters for variables instead of $f_i$, such as $t_i$.
For example, $(2.5f_1 + 3.1t_2 \geq 6) \wedge (f_1=1 \vee f_1=2) \wedge (f_1=2 \to t_2 \leq 1.1)$ is a formula by this definition. Observe that we allow standard abbreviations as $\neg (2.5f_1 + 3.1t_2 < 6)$ for $2.5f_1 + 3.1t_2 \geq 6$.

Since we are assuming the semantics of formulas over the domain of real numbers, an \textit{assignment} $\mathcal{A}$ for a formula $F$ is a mapping from the first-order variables of $F$ to elements in the domain of real numbers. For instance, $\{f_1 \mapsto 2.3, f_2 \mapsto 1\}$ is an assignment for $(2.5f_1 + 3.1f_2 \geq 6) \wedge (f_1=1 \vee f_1=2) \wedge (f_1=2 \to f_2 \leq 1.1)$. An assignment $\mathcal{A}$ \textit{satisfies} a formula $F$ if $F$ is true under this assignment. For example, $\{f_1 \mapsto 2, f_2 \mapsto 1.05\}$ satisfies the formula in the above example, whereas $\{f_1 \mapsto 2.3, f_2 \mapsto 1\}$ does not satisfy it. Moreover, an assignment $\mathcal{A}$ \textit{satisfies} a set $\Gamma$ of formulas if all formulas in $\Gamma$ are true under $\mathcal{A}$. A set of formulas $\Gamma$ is \textit{satisfiable} if there exists a satisfying assignment for $\Gamma$. To give an example, the set $\{(2.5f_1 + 3.1f_2 \geq 6), (f_1=1 \vee f_1=2), (f_1=2 \to f_2 \leq 1.1)\}$ is satisfiable since $\{f_1 \mapsto 2, f_2 \mapsto 1.05\}$ satisfies it. As another example, the set $\{(f_1 \geq 2), (f_1 < 1)\}$ is unsatisfiable since no assignment satisfies it.

\subsection{Logic-Based Explanations}\label{logic-based-exps}

Explanation models based on computational logic \cite{Xreason2022}, analyze the internal structure of machine learning algorithms to provide interpretable and correct explanations. Given an instance, logic-based methods identify a subset of input features sufficient to justify the correspondent output given by the model. Moreover, these methods ensure that explanations do not include any redundant features. In other words, removing any feature from the explanation would result in a subset that no longer guarantees the same prediction.

%By focusing on tree-based algorithms like XGBoost \cite{jemaa2024extending}, which uses regression tree structures, these models can be represented as a set of logical rules. These rules describe how the model generates predictions. 

The logic-based approaches \cite{ignatiev2019validating} to obtain correct and non-redundant explanations for XGBoost works as follows. First, given a specific instance we want to explain, we represent it by a conjunction of variables representing features and its corresponding values in the input:

\begin{equation}
    \mathcal{I} =  \bigwedge_{k=1}^n (f_k = v_k)
    \label{eq:iexp}
\end{equation}

where $f_k$ is a variable representing the $k$-th feature, $v_k$ is the specific value of the $k$-th feature in the instance, and $n$ is the number of features.

In the second step, each regression tree is encoded as a set of formulas $\mathcal{T}$. Each path in each regression tree of XGBoost is represented as an implication. For an example, see the set of formulas below that describe the regression tree in Figure~\ref{fig:decision_tree}. The variables in this formula are: age, income, and $o_1$. The variable $o_1$ represents the output of the first regression tree.

\begin{equation}
\label{eq:example_texp}
\mathcal{T} = 
\begin{cases}
\text{age} \geq 30 \land \text{income} \geq 50k \rightarrow o_1 = -1 \\
\text{age} \geq 30 \land \text{income} < 50k \rightarrow o_1 = 1 \\
\text{age} < 30 \land \text{age} \geq 25 \rightarrow o_1 = 3 \\
\text{age} < 30 \land \text{age} < 25 \rightarrow o_1 = -2
\end{cases}
\end{equation}

In a more general way:

\begin{equation}
    \mathcal{E} :=  \bigwedge_{h = 1}^m \left[ \bigwedge_{P \in R_h}\left(\bigwedge_{f_j \odot l_{P, j} \in P} (f_j \odot l_{P, j}) \rightarrow o_h = l_{P}\right) \right]
    \label{eq:texp}
\end{equation}

where $h$ is the regression tree index with $m$ being the number of trees, each $P$ is a path in regression tree of index $h$, i.e., $R_h$, and each $f_j$ is a feature occurring in path $P$. Moreover, $\odot$ is the comparison operator ($\geq$, $<$), $l_{P, j}$ is the comparison threshold for feature $f_j$ in path $P$, $o_h$ is the variable representing the output of regression tree $h$, and $l_p$ is the output value of the path $P$.

In the next step, the logic-based approach consists of defining a formula $\mathcal{D}$ that captures the prediction of the XGBoost. The final prediction is the sum of the outputs of all trees, where a positive sum indicates class 1 and a negative sum indicates class 0. Then, the prediction of the model is obtained by combining the outputs of all regression trees in the following way:

\begin{equation}
    \mathcal{D} = 
    \begin{cases} 
        \sum_{h=1}^m o_h + \text{init} > 0, & \text{if class} = 1 \\
        \sum_{h=1}^m o_h + \text{init} < 0, & \text{if class} = 0
    \end{cases}
    \label{eq:dexp}
\end{equation}

where \text{init} is an initial value set by the model during training. To generate explanations for a given prediction, logic-based approaches use an iterative algorithm. An explanation is computed removing feature by feature from $\mathcal{I}$. In each iteration, we analyze the satisfiability of the formula 
\begin{equation}\label{eq:explainer}
    \mathcal{I} \wedge \mathcal{E} \wedge \neg \mathcal{D}.
\end{equation}
This formula represents the following question: is it possible to find an instance that follows the model ($\mathcal{E}$), and has a different classification than the original ($\neg \mathcal{D}$) if we remove some features from the initial explanation $\mathcal{I}$, allowing those features to take any value?

This process is made one feature at a time, meaning that the last removed feature can be excluded from the explanation if the solver could not find a combination of values for the removed features that change the predicted class, meaning that they do not impact the decision. Otherwise, if it was possible to change the class, the feature is considered important and is retained. The final explanation is composed of all features that, when removed, could change the classification outcome.

%In binary classification, XGBoost constructs an ensemble of regression trees during training, each producing output values interpreted as probabilities. The final prediction is the sum of the outputs of all trees, where a positive sum indicates class 1 and a negative sum indicates class 0. 

%The ability to interpret the regression trees in a gradient boosting model allows for their mathematical representation, which forms the basis for developing explanation methods grounded in computational logic \cite{ignatiev2019validating}.

\section{Methodology}

This section details the methodology employed to develop a predictive model capable of identifying patients at risk of sudden death due to Chagas disease. We also describe the methodology used to apply logic-based explanations in this context, providing guarantees of correctness and enhancing the transparency of model predictions, thereby supporting clinical trust and decision-making.

\subsection{Data Collection Process}

The study utilized data from the Clementino Fraga Filho University Hospital at the Federal University of Rio de Janeiro, covering the period from 1992 to 2023 and consisting of clinical examinations and medical attributes from hospital patients. This forms a dataset of 452 exams with 252 patients, with various medical and health-related data, and categorizes each case in both general death cases and sudden death due to Chagas disease.

To build the final dataset for our application, the latest exams of each patient were used. Patients with severe cardiac impairment were excluded, as they were already candidates for interventions to ensure survival. After this exclusion, 120 patients remained for analysis, with 49 original features (30 binary, 12 continuous, and 7 categorical), with 19 patients affected by sudden death. These features were derived from cardiac tests, including echocardiograms, electrocardiograms, and 24-hour Holter ECGs. Due to the specific characteristics of the data, the relatively low number of samples should not be an issue for the model learning, as they capture sufficient information for this specific problem. The reduced number of instances is due to Chagas disease (CD) being a Neglected Tropical Disease (NTD) and the lack of large-scale databases available for early-stage research.
% explicar melhor o risco excluido
%For these reasons as well, the generation of synthetic data was excluded.

\subsection{Machine Learning Process}

\subsubsection{Feature Selection}

Since the model is intended to support patient classification, reducing the number of required features is essential to minimize the need for multiple medical exams, thus reducing costs and time requirements. Furthermore, since the data includes a wide range of general medical tests, certain features may exhibit low or negligible relevance to the target outcome. To address these concerns, feature selection is incorporated as part of the model training process to eliminate irrelevant or weakly correlated input features.

As a preparation step, categorical features were encoded using one-hot encoding. This transformation converts each categorical feature into binary features. After preprocessing, the dataset contains a total of 63 features. This number will be used for all subsequent analyses, including feature selection procedures.

\begin{table}[ht]
\centering
\caption{The 63 features, derived from one-hot encoding of 49 original input features, grouped according to their importance scores.}
\label{tab:grouped_features}
\begin{tabular}{|P{1.65cm}|c|P{9cm}|}
\hline
\textbf{Group} & \textbf{Qnt.} & \textbf{Features} \\
\hline
Top 10 & 10 &
EVTotal; NSVT (mean duration); LVEDV; LA diameter; EF Teicholz; Rassi score (points); LVESV; TSH; Dyslipidemia (DLP); Amiodarone \\
\hline
Top 20 & 20 &
\textbf{Top 10} + Holter Age; Stroke.1; ESV; NYHA Class I; HR; Rassi score\_low; Sex; LVEF; Mean HR; Stroke \\
\hline
Positive Importance & 48 &
\textbf{Top 20} + Syncope; NYHA Class II; BMI; Diastolic Dysfunction\_1.0; Classification\_Moderate Dysfunction; Guideline 2005\_B1; Type 2 Diabetes (DM2); Diastolic Dysfunction\_2.0; Ventricular Ectopy (VE); Rassi score\_intermediate; Hypertension (HTN); Sedentary lifestyle; Guideline 2005\_B2; Coronary artery disease; Classification\_Normal; Primary Alteration; Inactive Electrical Area; VE.1; Classification\_Mild Dysfunction; ICD (Implantable Defibrillator); Heart Failure; Chronic Kidney Disease; NSVT; Segmental Deficit; AV Conduction Disturbance.1\_0.0; Pulmonary Venous Doppler (PVD); Atrial Fibrillation/Flutter; Sinoatrial Node Dysfunction \\
\hline
Zero Importance & 15 &
AFib/Atrial Flutter; Smoking; Pacemaker; Cancer; Pause > 3s; Guideline 2005\_A; Ablations; AV Conduction Disturbance.1\_1.0; AV Conduction Disturbance\_3.0; AV Conduction Disturbance\_2.0; AV Conduction Disturbance\_1.0; AV Conduction Disturbance\_0.0; Diastolic Dysfunction\_0.0; Diastolic Dysfunction\_3.0; NYHA Class III \\
\hline
\end{tabular}
\end{table}

\subsubsection{Model Selection}

The objective of the model is to achieve a good overall recall score and F1 score, as it minimizes the false negatives of the predictions, which is critical for models used for diseases and, in this case, sudden death prevention.

In order to build the model, both feature selection and hyperparameter optimization were employed. Firstly, the XGBoost classifier was trained and evaluated using $k$-fold cross-validation with 5 folds and the default hyperparameters. The results revealed that not all 63 features contributed meaningfully to the model. Using the built-in \emph{feature importance} metric, we observed that 20 features exhibited zero or negligible importance, indicating they could be excluded without compromising performance.

We then sorted each feature by its mean feature importance value and selected subsets of features grouped into four categories, as shown in Table~\ref{tab:grouped_features}. Features with an importance value of zero were excluded from the subsequent tests. We built one model using all features from the \textit{Positive Importance} group; then, we created another using only the best 20 features, followed by a final model that utilized the best 10 features.

Each group was used to train the model, and to prevent possible overfitting, a k-fold cross-validation method was applied. Five stratified folds were created to balance the proportions of positive and negative labels in each fold. The model was trained and evaluated accordingly. During this process, Optuna\footnote{https://github.com/optuna/optuna} was used to perform hyperparameter tuning, aiming to maximize the recall score of the model for each subset. The set of hyperparameters selected for training was: 
\begin{itemize}
    \item \texttt{max\_depth}: Integer values from 3 to 10.
    \item \texttt{min\_child\_weight}: Integer values from 1 to 10.
    \item \texttt{learning\_rate}: Float values from 0.01 to 0.3 (log-uniform distribution).
    \item \texttt{colsample\_bytree}: Float values from 0.5 to 1.0.
\end{itemize}

After performing the tests, the model with 20 features had the best overall performance, achieving the highest recall while using fewer features than the model with 48 features. In contrast, the model trained with only the best 10 features showed the weakest performance. Table~\ref{tab:metric_subsetskfold} shows results for the mean and standard deviation of the evaluation metrics for each group of features.

% \begin{table}[t]
% \centering
% \caption{Hyperparameter search space used for XGBoost during model tuning with Optuna.}
% \label{tab:xgb_hyperparams}
% \begin{tabular}{|l|l|}
% \hline
% \textbf{Hyperparameter}       & \textbf{Range or Distribution} \\
% \hline
% \texttt{n\_estimators}        & 50 to 150 (int) \\
% \texttt{max\_depth}           & 3 to 10 (int) \\
% \texttt{min\_child\_weight}   & 1 to 10 (int) \\
% \texttt{learning\_rate}       & 0.01 to 0.3 (float, log-uniform) \\
% \texttt{colsample\_bytree}    & 0.5 to 1.0 (float) \\
% % \texttt{subsample}            & 0.5 to 1.0 (float) \\
% % \texttt{gamma}                & 0 to 5 (float) \\
% % \texttt{reg\_alpha}           & $10^{-5}$ to $10^{-1}$ (float, log-uniform) \\
% % \texttt{reg\_lambda}          & $10^{-5}$ to $10^{-1}$ (float, log-uniform) \\
% % \texttt{scale\_pos\_weight}   & 4 to 6 (float) \\
% \hline
% \end{tabular}
% \end{table}

\begin{table}[ht]
\caption{Metric Results for Each Subset of Features using Optuna with $k$-fold}
\centering
\label{tab:metric_subsetskfold}
\begin{tabular}{|l|c|c|c|c|c|}
\hline
\textbf{Features} & \textbf{Accuracy} & \textbf{Recall} & \textbf{Precision} & \textbf{F1-score} & \textbf{ROC AUC} \\
\hline
All (63) & 0.92 $\pm$ 0.04 & 0.85 $\pm$ 0.14 & 0.74 $\pm$ 0.16 & 0.78 $\pm$ 0.13 & 0.89 $\pm$ 0.08 \\
Positive (48) & 0.94 $\pm$ 0.037 & 0.90 $\pm$ 0.137 & 0.78 $\pm$ 0.14 & 0.83 $\pm$ 0.11 & 0.92 $\pm$ 0.07 \\
\textbf{Top 20} & \textbf{0.95 $\pm$ 0.03} & \textbf{0.95 $\pm$ 0.11} & \textbf{0.81 $\pm$ 0.18} & \textbf{0.86 $\pm$ 0.09} & \textbf{0.95 $\pm$ 0.05} \\
Top 10 & 0.92 $\pm$ 0.05 & 0.78 $\pm$ 0.22 & 0.74 $\pm$ 0.18 & 0.76 $\pm$ 0.19 & 0.87 $\pm$ 0.12 \\
\hline
\end{tabular}
\end{table}

\subsection{Generating Explanations}
\label{sec:experiments}

%For tree-based models like XGBoost, logic-based explanation methods offer stronger guarantees compared to heuristic approaches such as LIME or Anchor. 
%Unlike these heuristic explanation techniques, logic-based methods can provide correct explanations, offering a higher level of reliability \cite{ignatiev2019validating}. 

After training a machine learning model, the next step is to generate explanations that provide additional insights for users or domain experts. These explanations enhance the interpretability of the model, allowing users to validate its predictions and build trust in its decision-making process.

In our approach, we use logic-based explanations since they offer stronger guarantees of correctness compared to heuristic approaches such as LIME or Anchors. As logic-based methods can provide correct explanations, they offer a higher level of reliability \cite{ignatiev2019validating}. 

An example of a logic-based method specifically designed to generate correct explanations for models trained with XGBoost is XReason \cite{Xreason2022}. However, the publicly\footnote{https://github.com/alexeyignatiev/xreason} available implementation of XReason is tightly coupled with an older version of XGBoost, which may lead to significant variations in model behavior and results.

To overcome this limitation, we developed an alternative logic-based XAI approach that leverages the Z3 solver\footnote{https://github.com/Z3Prover/z3}, an SMT (Satisfiability Modulo Theories) solver~\cite{de2008z3}, together with XGBoost (version 2.1.4). By accessing the trained regression trees through the available methods in XGBoost, our approach is able to generate explanations with correctness guarantees without relying on outdated software dependencies. The logical formulas are constructed following the principles outlined in earlier work \cite{ignatiev2019validating, Xreason2022, jemaa2024extending} and Section~\ref{logic-based-exps}.

Since the model trained with the Top 20 features obtained the best performance, we trained a final model with the Top 20 features using the entire dataset and the best configuration of hyperparameters. It is important to note that, although the final model is trained on the complete dataset, all performance metrics and tuning decisions were conducted using appropriate validation techniques to avoid overfitting and ensure the reliability of the interpretations.

The final model trained with 20 features on all datasets incorporates all available information and can be used as a reference for generating explanations. Our experiments proceed by extracting the regression tree information of the model and implementing the explainer. After that, explanations will be generated for every sample in our dataset. In total, 120 explanations were created, with 101 for class 0 and 19 for class 1.

\textbf{Evaluating Explanations.} To evaluate explanation fidelity, we generated a group of randomized synthetic samples for each explanation. These samples preserved the values specified in the explanation while varying other features. Each sample was submitted to the model, and fidelity was measured as the percentage of synthetic samples classified identically to the original instance used to generate the explanation.

For comparison, we applied the same procedure to explanations generated using the Anchors and LIME methods. We report metrics on explanation size, fidelity, and running time in Section~\ref{sec:result}, including mean and standard deviation, grouped by predicted class. The LIME method requires a fixed explanation size to generate explanations, given by the user. Default parameters were used alongside the value of 10 features as a size, as it is a number greater than our maximum explanation size generated using both the logic-based method and Anchors, as the main comparison point is explanation fidelity. All experiments were executed with a 13th Gen Intel(R) Core(TM) i7-1360P 2.20 GHz, 16 GB of RAM, and running Windows 11. 
%To construct and manipulate logical expressions to build explanations, we use tools such as NumPy for numerical operations \cite{oliphant2006guide}, Pandas for data processing \cite{mckinney2010data}, and Scikit-learn~\cite{pedregosa2011scikit} for data sampling and metric evaluations, while XGBoost~\cite{chen2016xgboost} serves as the main model. Finally, Optuna~\cite{akiba2019optuna} is applied for model hyperparameter optimization.

\section{Results}
\label{sec:result}

This section presents the results obtained from the application of logic-based explanations to the predictive model described previously.

% \subsection{Model Results}
\subsection{Building the Explainer}
Based on the description in \ref{sec:experiments}, we have defined a logic-based explainer for XGBoost models. From the model, the regression trees were accessed in the form of a DataFrame \footnote{https://pandas.pydata.org} using the method \textit{booster} from the XGBoost library. Using this, the formulas in \eqref{eq:texp} were recursively created by accessing the nodes and leaves of the tree represented in the DataFrame. 
% After this, the maximum and minimum values from the dataset were used to define \eqref{eq:cexp}.

From the explainer, an instance can be passed to generate the explanation. For this, the instance is split into features that will form the formula in \eqref{eq:iexp}. Finally, the model is used to check the actual prediction, and the value of \textit{init} is obtained, forming the formula in \eqref{eq:dexp}. For computing explanations, we use Z3 with the formula in \eqref{eq:explainer}. 

Due to the iterative nature of the explanation generation process—which involves removing features—we leverage the \textit{feature importance} of XGBoost to guide this process efficiently. Specifically, features with zero importance (i.e., those that never appear in any decision node of the regression trees) are identified and excluded early, as they do not influence the predictions. Additionally, all features are sorted in ascending order of importance, allowing the explainer to prioritize the removal of less influential features in the initial iterations. These optimizations improve both the runtime and the minimality of the resulting explanations.

\begin{table}[b]
\centering
\caption{Comparison of Lime, Anchors and Logic-Based Explanations with respect to Fidelity}
\label{tab:explainers_comparison}
\begin{tabular}{|l|c|c|c|c|c|c|}
\hline
\textbf{Method} & \textbf{Class} & \textbf{Samples} & \textbf{Synthetic} & \textbf{Exp. Size} & \textbf{Fidelity(\%)} & \textbf{Time(s)} \\
\hline
\multirow{2}{*}{Anchors} & 0 & 101 & 100 & $1.99 \pm 0.36$ & $84.68 \pm 15.32$ & $0.88 \pm 0.24$ \\
                         & 1 & 19  & 100 & $3.58 \pm 0.44$ & $74.26 \pm 25.74$ & $0.88 \pm 0.20$  \\
\hline
\multirow{2}{*}{Lime} & 0 & 101 & 100 & $10 \pm 0$ & $98.78 \pm 11.00$ & $0.05 \pm 0.01$ \\  
                         & 1 & 19  & 100 & $10 \pm 0$ & $97.79 \pm 14.71$ & $0.06 \pm 0.01$ \\
\hline
\multirow{2}{*}{Logic-based}   & 0 & 101 & 100 & $6.33 \pm 0.69$ & $\mathbf{100.00 \pm 0.00}$ & $0.38 \pm 0.11$ \\
                         & 1 & 19  & 100 & $3.89 \pm 1.52$ & $\mathbf{100.00 \pm 0.00}$ & $0.32 \pm 0.09$ \\
\hline
\end{tabular}
\end{table}

\subsection{Comparing Explanations}

To evaluate explanation fidelity, we generated synthetic samples that preserve the values of explanation features while randomizing the others, as explained in Section~\ref{sec:experiments}. These samples are expected to maintain the original classification, which is guaranteed by our logic-based explanations.

Table~\ref{tab:explainers_comparison} shows a comparison of our explanations' fidelity through synthetic samples compared to explanations from Anchors and Lime. Both are applied to the same XGBoost model and separated by classes 0 and 1.
According to Table \ref{tab:explainers_comparison}, the logic-based method consistently achieves 100\% fidelity in its explanations. This indicates a higher level of reliability when compared to the Anchors and LIME fidelity results. The logic-based explanations tend to have larger sizes when compared to the Anchors method, especially for class 0. This trade-off in explanation size is acceptable given the substantial gain in fidelity, making the logic-based method more trustworthy for applications where correctness is crucial, such as clinical decision support.

As for the LIME method, it uses a fixed explanation size when generating results. Even though we used an explanation size that was larger than the average size of our logic-based method, LIME could not achieve 100\% fidelity for its explanations. This limitation highlights the probabilistic nature of LIME’s sampling approach, which does not always capture the model's true decision boundaries, despite including more features.

In terms of time to generate explanations, LIME was the fastest method, while Anchors required more time than both methods tested. The logic-based method had intermediate performance. While not as fast as LIME, the logic-based method provides a balanced performance.

When generating explanations for this dataset using the logic-based method, the maximum size observed was 9 features for a sample of class 0 and 8 features for a sample of class 1. On the other hand, the minimum size observed was 5 and 3 for classes 0 and 1, respectively. This shows an interesting result, as explanations of class 0 were overall larger than class 1.

% In conclusion, while LIME and Anchors offer both faster and more concise explanations, respectively. Their lower fidelity can impact their trustworthiness in sensitive domains. The logic-based approach stands out by providing both interpretability and correctness, making it the most reliable and suitable method for high-stakes applications.

% \subsection{Explanation Sizes}

% Another important metric for explainable models is the size of the explanation, defined as the number of features returned by the method for a given sample. To evaluate this, we generated an explanation for each sample and used the resulting features as constraints to filter the dataset, grouping samples that satisfy the same explanation. By construction, each group is guaranteed to have a consistent classification across all its members. While it is uncommon for multiple samples to share exactly the same explanation, this grouping allows us to verify the model’s local consistency. In our dataset, we were able to generate 120 explanation-based groups, one per sample, which were then categorized by predicted class, as shown in Table~\ref{tab:explain_sizes}.

% \begin{table}[b]
% \centering
% \caption{Logic-Based Explanation Maximum, Minimum and Mean sizes}
% \label{tab:explain_sizes}
% \begin{tabular}{|l|c|c|c|c|c|c|}
% \hline
% \textbf{Class} & \textbf{Samples} & \textbf{Features} & \textbf{Explanation Size} & \textbf{Maximum Size} & \textbf{Minimum Size} \\
% \hline
% 0 & 101 & 20 &  $6.33 \pm 0.69$ & 9 & 5  \\
% 1 & 19  & 20 &  $3.89 \pm 1.52$ & 8 & 3 \\
% \hline
% \end{tabular}
% \end{table}

\subsection{Features in Explanations}

To explore the model's interpretability, we analyzed explanations generated for all samples in the dataset, focusing on the features used. Table~\ref{tab:explanation_examples} presents four illustrative examples produced by the logic-based method. These examples highlight the method’s ability to identify minimal subsets of features that justify the model’s classifications. Such explanations provide a more transparent understanding of the model’s decision-making process and reveal which key feature sets were important for each classification, allowing medical specialists to explore and validate them.

Moreover, for each sample, we checked the frequency by counting how many times a feature appeared in explanations for our data. These frequencies were then compared with the global feature importances derived from the XGBoost model. Table \ref{tab:feature_explanation_frequency} presents the comparison between explanation frequency and feature importance.

\begin{table}[t]
\centering
\caption{Examples of Explanations to Four Samples of the Chagas Disease Dataset}
\label{tab:explanation_examples}
\begin{tabular}{|c|c|p{9cm}|}
\hline
\textbf{Sample} & \textbf{Class} & \textbf{Explanation} \\
\hline
1 & High risk & LA diameter = 4.5, LVEDV = 5.8, EVTotal = 11094\\
2 & High risk & LVEDV = 5.8, EVTotal = 13812, NSVT (mean duration) = 1 \\
3 & Low risk  & LVEDV = 4.4, EF Teicholz = 0.64, Rassi score (points) = 3, Stroke = 0,
EVTotal = 1589, NSVT (mean duration) = 0\\
4 & Low risk  & LVEDV = 4.3, EF Teicholz = 0.68, Rassi score (points) = 2, Stroke = 0,
EVTotal = 236, NSVT (mean duration) = 0
 \\
\hline
\end{tabular}
\end{table}

\begin{table}[h]
\centering
\caption{Comparison Between Feature Presence Count in Explanations and Feature Importance, Ordered by Count.}
\label{tab:feature_explanation_frequency}
\begin{tabular}{|c|l|c|c|}
\hline

\textbf{Rank} & \textbf{Feature} & \textbf{Count in Explanations} & \textbf{Feature Importance} \\
\hline

1 & EVTotal        & 119 & 0.2488 \\
2 & LVEDV          & 114 & 0.0413 \\
3 & NSVT Holter    & 113 & 0.2756 \\
4 & Stroke         & 103 & 0.0608 \\
5 & Rassi Score(points) & 97  & 0.0492 \\
6 & TSH            & 87  & 0.0488 \\
7 & LVESV          & 26  & 0.0478 \\
8 & EF Teicholz    & 21  & 0.0471 \\
9 & LA Diameter    & 14  & 0.0399 \\
10 & Stroke.1      & 8   & 0.0310 \\
11 & ESV           & 6   & 0.0212 \\
12 & FEVE          & 3   & 0.0465 \\
13 & DLP           & 1   & 0.0138 \\
14 & Holter Age    & 0   & 0.0121 \\
15 & NYHA\_2.0     & 0   & 0.0098 \\
16 & HR            & 0   & 0.0062 \\
\hline
\end{tabular}
\end{table}

As shown in Table~\ref{tab:feature_explanation_frequency}, feature importance scores do not always align with how frequently features appear in explanations. For instance, although \textit{LVEDV} and \textit{Stroke} received relatively low importance scores, they appeared in a large number of explanations. Conversely, some features with non-zero importance scores were not included in any explanation for this dataset, suggesting that their influence is more marginal or redundant when explanations are constructed logically.
%\begin{itemize}
 %   \item \textbf{Instance 1} (class 1 – high risk)\\
 %   LA diameter = 4.5, LVEDV = 5.8, EVTotal = 11094

%    \item \textbf{Instance 2} (class 1 – high risk)\\
%    LVEDV = 5.8, EVTotal = 13812, NSVT (mean duration) = 1

%    \item \textbf{Instance 3} (class 0 – low risk)\\
 %   LVEDV = 4.4, EF Teicholz = 0.64, Rassi score (points) = 3, Stroke = 0, EVTotal = 1589, NSVT (mean duration) = 0

%    \item \textbf{Instance 4} (class 0 – low risk)\\
%    LVEDV = 4.3, EF Teicholz = 0.68, Rassi score (points) = 2, Stroke = 0, EVTotal = 236, NSVT (mean duration) = 0
%\end{itemize}

\section{Conclusions and Future Work}

This study focused on building a predictive model with interpretability, demonstrating how explainable machine learning can be applied in sensitive domains such as sudden death classification for Chagas disease. Beyond achieving good predictive performance, we focused on understanding and demonstrating how the model makes decisions and why explanations with correctness guarantees are important for the trust and transparency of the machine learning model.

%By implementing the explanations based on instances and comparing them to the explanations with ranges of features, we could showcase how grouping instances based on shared influential features can be insightful for understanding the model. This demonstrates how explanations can not only clarify predictions but also reveal meaningful patterns within how the model perceives the data.

Future approaches may explore adapting this model to multimodal scenarios and submitting the explanations for expert evaluation. Furthermore, exploration of how grouping of datasets could be made by using explanations and their application to the interpretability of the model.

\begin{credits}
\subsubsection*{\ackname}
The authors acknowledge the support of the Federal Institute of Education, Science and Technology of Ceará (IFCE) through the research grant calls PIBIC No. 11/2024 and 7/2024, issued by the PRPI/IFCE, as well as the support of Fundação Cearense de Apoio ao Desenvolvimento Científico e Tecnológico (FUNCAP) in the development of this work.

% A bold run-in heading in small font size at the end of the paper is
% used for general acknowledgments, for example: This study was funded

% \subsubsection{\discintname} ...
% It is now necessary to declare any competing interests or to specifically
% state that the authors have no competing interests. Please place the
% statement with a bold run-in heading in small font size beneath the
% (optional) acknowledgments\footnote{If EquinOCS, our proceedings submission
% system, is used, then the disclaimer can be provided directly in the system.},
% for example: The authors have no competing interests to declare that are
% relevant to the content of this article. Or: Author A has received research
% grants from Company W. Author B has received a speaker honorarium from
% Company X and owns stock in Company Y. Author C is a member of committee Z.
\end{credits}
\bibliographystyle{splncs04nat}
\bibliography{references}

% \end{thebibliography}
\end{document}